\pdfoutput=1

\documentclass[11pt]{article}

\usepackage{acl}

\usepackage{times}
\usepackage{latexsym}

\usepackage[T1]{fontenc}

\usepackage[utf8]{inputenc}

\usepackage{microtype}

\usepackage{refstyle}
\usepackage{booktabs}
\usepackage{multirow} %
\usepackage{soul}%
\usepackage{colortbl}

\usepackage{graphicx}

\usepackage{amsmath}

\usepackage{cleveref}
\crefformat{section}{\S#2#1#3}
\crefformat{subsection}{\S#2#1#3}
\crefformat{subsubsection}{\S#2#1#3}
\crefrangeformat{section}{\S#3#1#4 to~\S#5#2#6}
\crefmultiformat{section}{\S#2#1#3}{ and~\S#2#1#3}{, #2#1#3}{ and~#2#1#3}
\Crefformat{figure}{#2Fig.~#1#3}
\Crefmultiformat{figure}{Figs.~#2#1#3}{ and~#2#1#3}{, #2#1#3}{ and~#2#1#3}
\Crefformat{table}{#2Tab.~#1#3}
\Crefmultiformat{table}{Tabs.~#2#1#3}{ and~#2#1#3}{, #2#1#3}{ and~#2#1#3}
\Crefformat{appendix}{Appx.~\S#2#1#3}
\crefformat{algorithm}{Alg.~#2#1#3}

\newcommand{\stitle}[1]{\vspace{1ex} \noindent{\bf #1.}}

\usepackage{xspace}
\newcommand{\MODEL}{\mbox{\textsc{Salt}}\xspace} %

\title{Self-Augmentation Improves Zero-Shot Cross-Lingual Transfer}

\author{
  Fei Wang$^\dag$\;~\;~ Kuan-Hao Huang$^\ddag$\;~\;~ Kai-Wei Chang$^\ddag$\;~\;~ Muhao Chen$^\dag$ \\
  $^\dag$University of Southern California $\ $ $^\ddag$University of California, Los Angeles \\
  \{fwang598, muhaoche\}@usc.edu $\ $ \{khhuang, kwchang\}@cs.ucla.edu
  }

\begin{document}
\maketitle

\begin{abstract}

Zero-shot cross-lingual transfer is a central task in multilingual NLP, allowing models trained in languages with more sufficient training resources to generalize to other low-resource languages.
Earlier efforts on this task use parallel corpora, bilingual dictionaries, or other annotated alignment data to improve cross-lingual transferability, which are typically expensive to obtain.
In this paper, we propose a simple yet effective method, \MODEL, to improve the \emph{zero-shot} cross-lingual transfer of the multilingual pretrained language models without the help of such external data.
By incorporating code-switching and embedding mixup with self-augmentation,
\MODEL effectively distills cross-lingual knowledge from the multilingual PLM and enhances its transferability on downstream tasks.
Experimental results on XNLI and PAWS-X show that our method is able to improve zero-shot cross-lingual transferability without external data.\footnote{Our code is available at \url{https://github.com/luka-group/SALT}.}

\end{abstract}

\section{Introduction}

Zero-shot cross-lingual transfer %
is integral to many
multilingual NLP tasks \cite{ma2014unsupervised,artetxe2019massively,ahmad2019difficulties}.
For some NLP tasks, the
task-specific training data are %
often not evenly provided %
in terms of quantity and quality for distinct languages,
and may even be unavailable for %
particularly low-resource languages.
Zero-shot cross-lingual transfer allows models trained %
in languages with sufficient training resources
to generalize to other low-resource languages. %
Earlier efforts on this task use %
parallel corpora, bilingual dictionaries, or other annotated alignment data to improve cross-lingual transferability, which %
are typically expensive to obtain \cite{Chi21trans2,Yang22tran1,qin2020cosda,lee2021scopa,krishnan2021multilingual}.

Recent analyses show that multilingual pre-trained language models (PLMs) possess rich cross-lingual knowledge to facilitate the transfer \cite{pires2019multilingual,conneau2020unsupervised,huang2021improving}.
However, when being finetuned on a specific task in the source language,
multilingual PLMs %
may catastrophically forget cross-lingual knowledge
\cite{liu2021preserving,chalkidis2021multieurlex}.
We argue that cross-lingual knowledge possessed by multilingual PLMs can be distilled and incorporated into %
task training data to preserve and improve models' cross-lingual transferability %
when fine-tuning on downstream tasks.

\begin{figure}[t]
      \begin{center}
        \includegraphics[width=\columnwidth]{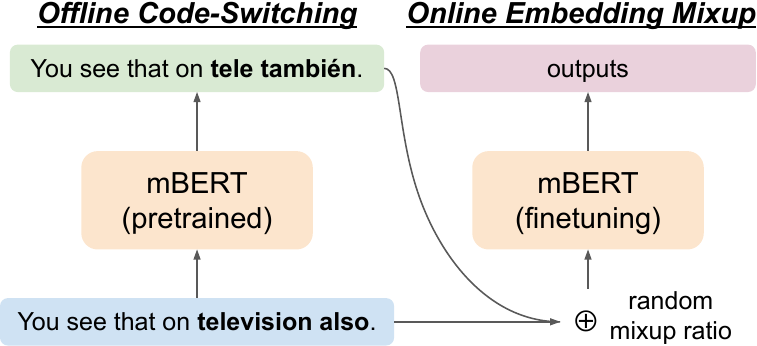} %
      \end{center}
      \caption{\MODEL distills the cross-lingual counterpart of each token from the multilingual PLM for code-switching and applies embedding mixup to improve the diversity of self-augmented features. %
      }
      \label{fig:model}
    \end{figure}

In this paper, we focus on the setting where no external cross-lingual alignment data are available and propose a simple yet effective Self-Augmented Language Transfer (\MODEL) method for multilingual PLMs.
\MODEL introduces two self-augmentation techniques on monolingual task-specific training data as shown in \Cref{fig:model}.
Before task training, an \emph{offline} technique based on cross-lingual code-switching
first uses the PLM to predict a cross-lingual counterpart of each token in the given %
training sample.
Accordingly, this technique generates a cross-lingual augmentation of the training sample by substituting the tokens with their cross-lingual counterparts.
During task training, an \emph{online} self-augmentation technique based on embedding mixup \cite{zhang2018mixup} randomly perturbs the representation of the training sample between the source and target languages.
\MODEL allows the information about context-specific alignment of tokens to be distilled from the multilingual PLM,
and to be further enhanced through perturbed training.

Experimental results on XNLI and PAWS-X benchmarks demonstrate that \MODEL %
achieves more improvements on zero-shot transferability than previous SOTA methods \cite{Lewis20generalized,huang2021improving}.
With self-augmentation on three target languages, \MODEL achieves 1.4\% improvement on 15 languages of XNLI and 4.1\% on 6 languages of PAWS-X in terms of average accuracy comparing with the base model.

\section{Self-Augmentation}

Our method distills cross-lingual token-level alignment from the multilingual PLM and incorporates the knowledge to task-specific data through code-switching. To further improve the diversity of augmentation for better generalization, we also apply a random embedding mixup.
Following are the details of these two techniques.

\subsection{Self-Augmentation with Code-Switching}

\MODEL adopts a modified masked language modeling (MLM) method to distill cross-lingual token translation pairs. 
As multilingual PLMs, such as mBERT \cite{Devlin19bert} and XLM-R \cite{conneau2020unsupervised}, are pre-trained with MLM, we can directly %
mask these models to predict cross-lingual tokens.
Specifically, \MODEL makes two modifications to the original MLM learning objective. 
First, we let the model predict tokens only in a specific target language, and disregard all tokens not belonging to this language. 
Tokens in each target language are collected from a monolingual vocabulary list.\footnote{We adopt the top 10,000 words by frequency in target languages at \url{https://en.wiktionary.org/wiki/Wiktionary:Frequency_lists}.} 
Second, to ensure that the semantics of predictions are similar to original tokens, we do not mask the original tokens %
when inputting the sentence to the multilingual PLM.
This seeks to help the PLM better infer cross-lingual counterpart tokens given the references of original tokens in the input context.

Generated token-level cross-lingual substitutions with high enough predicted probabilities are used for code-switching.
In this work, we adopt a fixed probability threshold for all target languages.
We also predict synonyms in the source language with \MODEL.
Since predicted probability in the source language and target languages are of different scales, we assign a different threshold for synonym prediction. 
This self-augmentation process is done offline before task training.

\subsection{Self-Augmentation with Embedding Mixup}

To further improve the diversity of self-augmented features and reduce the noise introduced by code-switching through smoothing, we propose an \emph{online} self-augmentation technique based on
cross-lingual embedding mixup \cite{guo2019augmenting}.
Specifically, for each token $t_i$ in the generated sentence from code-switching, we interpolate its embedding with that of the original token $s_i$ before code-switching.
In detail,
given the embedding of original token $h_{s_i}$ and embedding of the substituted token $h_{t_i}$, 
the mixed up token embedding is generated as
$$h_i = r \cdot h_{s_i} + (1-r) \cdot h_{t_i},$$
where $r=\{r_{j}\}, r_{j} \in [0,1]$ is a random vector. 
In the embedding mixup, the interpolation coefficient of each embedding dimension $j$ is independently generated from a uniform distribution to allow for more diverse combinations of embeddings in two languages.
Applying this self-augmentation technique allows for randomly perturbing the training instance between the source and target language representation, leading to further improved transferability of the obtained task model.

\begin{table*}[t]
\centering
\small
\setlength{\tabcolsep}{3.5pt}

\begin{tabular}{lc|cccccccccccccc|cc}
    \toprule
    Model             & en   & ar   & bg   & de   & el   & es   & fr   & hi   & ru   & sw   & th   & tr   & ur   & vi   & zh & avg.  & w/ en \\
    \midrule
    \multicolumn{18}{l}{\cellcolor{blue!10}\textit{without external data}} \\
    \midrule
    mBERT             & 80.8 & 64.3 & 68.0 & 70.0 & 65.3 & 73.5 & 73.4 & 58.9 & 67.8 & 49.7 & \textbf{54.1} & 60.9 & 57.2 & 69.3 & 67.8 & 64.3 & 65.4 \\
    mBERT*            & 81.8 & 64.7	& 67.0 & 69.6 & 66.1 & 74.7 & 73.9 & 59.7 & 68.8 & 50.0 & 53.7 & 60.8 & 58.0 & 70.5 & 68.8 & 64.7 & 66.0 \\
    +RS-ADV         & 81.9 & 64.9 & 68.3 & \textbf{71.7} & 66.5 & 74.4 & \textbf{74.5} & 59.6 & 68.8 & 48.8 & 50.6 & 61.7 & \textbf{59.2} & 70.0 & 69.4 & 64.9 & 66.0 \\
    +RS-RP       & \textbf{82.6} & 65.4 & 68.7 & 70.5 & 67.2 & \textbf{75.0} & 74.1 & 59.8 & 69.5 & 48.4 & 50.5 & 59.7 & 57.9 & 70.5 & 69.7 & 64.8 & 66.0 \\ \midrule
    +\MODEL     & \underline{82.4} & \underline{\textbf{65.9}}	& \underline{\textbf{69.6}} & 70.4 & \underline{\textbf{67.8}} & \textbf{75.0} & \textbf{74.5} & \underline{\textbf{61.1}} & \textbf{69.6} & \underline{\textbf{50.6}}	& 53.5 & \underline{\textbf{62.0}} & \underline{58.9} & \textbf{70.9}	& \underline{\textbf{69.8}} & \underline{\textbf{65.7}} & \underline{\textbf{66.8}} \\
    \midrule
    \multicolumn{18}{l}{\cellcolor{blue!10}\textit{with external data (not directly comparable to our approach)}} \\
    \midrule
    +RS-DA       & 81.0 & 66.4 & 69.9 & 71.8 & 68.0 & 74.7 & 74.2 & 62.7 & 70.6 & 51.1 & 55.7 & 62.9 & 60.9 & 71.8 & 71.4 & 66.6 & 67.6 \\
    +CoSDA-ML    & 82.9 & 68.0 & 72.7 & 74.1 & 70.9 & 76.9 & 76.7 & 65.5 & 73.2 & 51.0 & 59.8 & 63.9 & 62.3 & 73.6 & 73.8 & 68.7 & 69.7 \\
    \bottomrule
\end{tabular}
\caption{Average accuracy of zero-shot cross-lingual transfer on XNLI with 5 different random seeds. We provide detailed results of 15 languages and their average (w/ and w/o English). The highest scores are in bold. Significant improvements in comparison with mBERT baseline by t-test ($p \leq 0.05$) are underlined. We also provide results of methods with external data as the referenced upper bound. * We reproduce mBERT with our code base. Other baseline results are from previous papers.}
\label{tab:xnli}
\end{table*}

\subsection{Training}
In this work, we use English as the source language and consider three target languages for code-switching, i.e. French, German and Spanish\footnote{We compute the overlap between the model's vocabulary and the word list of each evaluation language, and select the top three languages with the highest overlap ratio. Note that Chinese, Japanese, and Korean are not selected because their words will be tokenized to characters by the model tokenizer used in this study.}.
Before training the task model,
on each original training sample in the source language, we generate an augmented sample in each target language \emph{offline} with code-switching, where all predicted high-probability token substitutions are applied.
Then during training, we %
further apply embedding mixup to these augmented samples
where the interpolation coefficient vector $r$ is dynamically sampled in each step of training for each instance. %
While both self-augmentation techniques automatically switch and perturb the original training samples towards the target language(s), the labels on those training samples remain unchanged.
Hence, the final task model is trained directly on the self-augmented training samples by optimizing the original learning objective of the task.
This allows for robust zero-shot transfer of the task model by using only monolingual training data in the source language.

\section{Experiment}

\begin{table*}[t]
\centering
\small
\setlength{\tabcolsep}{7pt}
\begin{tabular}{lc|cccccc|cc}
    \toprule
    Model       & en   & de   & es   & fr   & ja   & ko   & zh   & avg. & w/ en \\
    \midrule
    \multicolumn{10}{l}{\cellcolor{blue!10}\textit{without external data}} \\
    \midrule
    mBERT             & 94.0 & 85.7 & 87.4 & 87.0 & 73.0 & 69.6 & 77.0 & 80.0 & 82.0 \\
    mBERT*            & 93.9 & 84.5	& 87.9 & 87.2 & 74.2 & 76.1	& 79.8 & 81.6 & 83.4 \\
    +RS-ADV      & 93.7 & 86.5 & 88.5 & 87.8 & 76.1 & 75.3 & 80.4 & 82.4 & 84.0 \\
    +RS-RP       & \textbf{94.5} & 87.4 & \textbf{90.0} & \textbf{89.5} & 77.9 & \textbf{77.5} & \textbf{82.0} & \textbf{84.1} & \textbf{85.5} \\\midrule
    +\MODEL      & 94.2 & \textbf{87.9}	& 89.9 & 89.1 & \textbf{78.6} & 77.4 & 81.8 & \textbf{84.1} & \textbf{85.5} \\
    \midrule
    \multicolumn{10}{l}{\cellcolor{blue!10}\textit{with external data (not directly comparable to our approach)}} \\ 
    \midrule
    +RS-DA       & 93.5 & 87.8 & 88.8 & 88.8 & 79.3 & 78.3 & 81.5 & 84.1 & 85.4 \\
    +CoSDA-ML    & -    & 87.3 & 90.0 & 89.6 & 79.4 & 79.5 & 83.0 & 84.8 & -    \\
    +SCOPA       & -    & 88.6 & 90.3 & 89.7 & 81.5 & 80.1 & 84.1 & 85.7 & -    \\
    \bottomrule
\end{tabular}
\caption{Average accuracy of zero-shot cross-lingual transfer on PAWS-X. %
}
\label{tab:paws-x}
\end{table*}

In this section, we evaluate \MODEL to demonstrate that self-augmentation methods improve zero-shot cross-lingual transfer.

\subsection{Setup}
Following \citet{huang2021improving}, %
we consider two cross-lingual datasets.
XNLI is a natural language inference (NLI) dataset, including premise-hypothesis pairs in 15 languages labeled as entailment/neutral/contradiction. 
PAWS-X is a paraphrase identification dataset, including sentence pairs in six languages with binary labels.
On both datasets, we train and validate models with English data, and test them with data in all languages. We report the average accuracy of five-run experiment.

\stitle{Baseline}
We compare \MODEL with two SOTA zero-shot cross-lingual transfer methods without any external data.
RS-ADV and RS-RP \cite{huang2021improving} enhance transferability by respectively adding adversarial and random embedding perturbation %
during task training on English data. 
We also provide results of three methods that use external data.
RS-DA \cite{huang2021improving} augments training data by replacing words with predefined synonyms \cite{alzantot-etal-2018-generating}.
CoSDA-ML \cite{qin2020cosda} creates code-switching data with bilingual dictionaries.
SCOPA \cite{lee2021scopa} extends CoSDA-ML by mixing the hidden states of original and code-switched data with a fixed ratio.

\begin{table}[t]
\centering
\small
\setlength{\tabcolsep}{7pt}
\begin{tabular}{lcc}
    \toprule
    Model       &  avg. & avg. (incl. en) \\
    \midrule
    \MODEL      &  \textbf{84.1} & \textbf{85.5} \\
    - en-only & 83.4 &	84.8 \\
    - w/o mixup & 83.3 & 84.7 \\ \midrule
    mBERT & 81.6 & 83.4 \\
    \bottomrule
\end{tabular}
\caption{Ablation study on PAWS-X test set. \textit{en-only} indicates only substitute original tokens to other English tokens. \textit{w/o mixup} means embedding mixup is not used.}
\label{tab:ablation}
\end{table}

\stitle{Implementation Details}
We evaluate the proposed method based on mBERT.
For both sentence pair classification tasks, self-augmentation is conducted separately on each sentence. We augment one code-switched instance per language (including English, French, Spanish, and German) for each original instance. For other hyper-parameters, we follow the training scripts by \citet{huang2021improving}.  More details are in \Cref{sec:appendix_implementation}.

\subsection{Results}
\label{sec:results}
\Cref{tab:xnli} shows the results on XNLI, where we observe that previous methods without external data have a positive influence on a few languages but have a negative influence on other languages. %
Their average improvement is however lesser in comparison to \MODEL which leads to an average improvement in accuracy by 1\% as well as
significant improvements in 9 out of 15 languages over mBERT.
Our method also outperforms other baselines by at least 0.8\% in terms of average accuracy (both w/ and w/o en). 
The results indicate that the model can benefit from cross-lingual knowledge distilled from itself. Moreover, augmenting the data to three target languages can bring improvements to all 14 target languages.
For example, the improvements on ar, bg, el, hi, tr and zh are 1.2\%, 2.6\%, 1.7\%, 1.4\%, 1.2\% and 1.0\%.
Experiment on PAWS-X also shows that \MODEL can improve model performance in comparison with the vanilla setting (\Cref{tab:paws-x}). However, RS-RP is also effective on this task and achieves comparable results. %
Considering that NLI requires inferring the logical consequence that can be dependent on various components of the two sentences, this complex reasoning process benefits more from the robust training of \MODEL.
On the other hand, as a simpler task based on sentence similarity, paraphrase identification can be sufficiently improved based on random perturbation. 

\stitle{Generalized Setting}
We further evaluate \MODEL in a generalized setting \cite{Lewis20generalized,huang2021improving} on XNLI. The new setting pairs up sentences from two different languages as the premise and the hypothesis, converting the original test data from 15 languages to 225 language pairs. \MODEL achieves 0.5\% of improvement over the best baseline and 2.2\% over the vanilla PLM in average. Full results on all language pairs are in \Cref{sec:appendix_result}. Despite the vocabulary gap between training and inference for baselines, \MODEL reduces this gap by code-switching and mixup. 

\stitle{Ablation Study}
To %
further investigate the incorporated techniques
in \MODEL, we conduct an ablation study on PAWS-X as shown in \Cref{tab:ablation}. Offline code-switching solely improves the average accuracy by 1.7\%, while online embedding mixup further improves it by 0.8\%. %
We also evaluate the influence of involved target languages in \MODEL. Code-switching with only English synonyms distilled from PLMs can bring an improvement of 1.8\%, while involving three target languages further improves the performance by 0.7\%.

\section{Related Work}

Zero-shot cross-lingual transfer has become an emerging research topic since it potentially reduces the effort of collecting labeled data for low-resource languages \cite{ahmad2019difficulties,Hu20xtreme,Dufter20emerging1,Ruder21xtremer,ChaiLD22emerging3,Huang22xgear}.
Earlier works directly apply multilingual PLMs, such as multilingual BERT \cite{Devlin19bert}, XLM \cite{ConneauL19xlm}, and XLM-R \cite{conneau2020unsupervised}, and achieve surprisingly well performance on  this setting.
Recently, the performance is further improved with additional auxiliary data, such as parallel translation pairs \cite{Chi21trans2,Wei21tans4,Yang22tran1,Feng22trans3}, bilingual dictionaries \cite{Cao20align1,qin2020cosda,Liu20codesw2,krishnan2021multilingual,lee2021scopa}, and syntactic features  \cite{Subburathinam19clgcn,meng2019target,Ahmad21syntaxbert,Ahmad21gate}.

Our work aligns more with another line of research that studies zero-shot cross-lingual transfer \emph{without} using additional annotations.
This includes unsupervised embedding alignment \cite{Artetxe20ualign2,Conneau20ualign1}, robust training \cite{huang2021improving}, and meta-learning \cite{Nooralahzadeh20meta}.
Our idea is motivated by the self-augmentation techniques \cite{Feng21selfaug,xu2021sas} that are mostly explored for monolingual tasks, and the mixup techniques \cite{zhang2018mixup,lee2021scopa,Yang22tran1} which seeks to smooth the embedding space.

\section{Conclusion}

In this paper, we propose \MODEL, a self-augmention method for zero-shot cross-lingual transfer of PLMs. 
\MODEL distills cross-lingual knowledge from PLMs and incorporates them into task training data through an offline code-switching technique,
and an online embedding mixup technique to improve transferability with a smoothed representation space.
Experiments and analyses based on XNLI and PAWS-X demonstrate promising improvement to \MODEL in terms of cross-lingual transfer without using external data. 

\section*{Acknowledgement}
We appreciate the reviewers for their insightful comments and suggestions.
Fei Wang was supported by the Annenberg Fellowship at USC and the Amazon ML Fellowship.
Muhao Chen was supported by the NSF Grants IIS 2105329 and ITE 2333736, by the Air Force Research Laboratory under
agreement number FA8750-20-2-10002, by two Amazon Research Awards and a Cisco Research Award.
Computing of this work was partly supported by a subaward of NSF Cloudbank 1925001 through UCSD.

\section*{Limitations}
In this study, we adopt a fixed threshold to select tokens for code-switching. However, the optimal thresholds for different languages and instances can vary. Future research can develop efficient search algorithm to optimize the thresholds.
While we have limited the proposed technique to descriminative natural language understanding tasks, future research can extend the proposed technique to generative multilingual PLMs, such as mT5 \cite{xue2021mt5} and mBART \cite{liu2020multilingual}, on text generation tasks \cite{duan2019zero,chen2021zero}. 
Furthermore, we have opted for English as the source language. However, extending the application of SALT to other source languages could enhance the comprehensiveness of this study.

\bibliography{anthology,reference}
\bibliographystyle{acl_natbib}

\clearpage
\appendix

\section{Implementation Details}
\label{sec:appendix_implementation}
We implement our model based on Huggingface Transformers~\cite{wolf2019huggingface}. 
We apply the uncased base version of mBERT model consisting of 110M parameters.
We set the probability threshold for token substitution to 1e-3 for English synonym replacement and 1e-7 for code-switching in other languages based on the development set.
We run experiments with a NVIDIA GeForce RTX 2080 GPU. 
It takes about 1 hour for training on PAWS-X and 5 hours on XNLI.

\section{Results of Generalized Setting}
Results for \MODEL, RS-RP and vanilla mBERT on XNLI under generalized setting are shown in \Cref{tab:gen_salt}, \Cref{tab:gen_rsrp} and \Cref{tab:gen_mbert} respectively. Baseline results of RS-RP and vanilla mBERT are copied from \citet{huang2021improving}. 

\label{sec:appendix_result}
\begin{table*}[h]
    \centering
    \small
    \setlength{\tabcolsep}{3.5pt}
    \begin{tabular}{c|ccccccccccccccc|c}
    \toprule
        ~ & en & es & de & fr & bg & ru & el & th & sw & vi & ar & zh & hi & ur & tr & avg.  \\ \midrule
        en & 82.9 & 73.2 & 67.5 & 72.3 & 66.5 & 67.4 & 60.7 & 46.6 & 40.4 & 67.4 & 62.1 & 68.7 & 56.9 & 53.6 & 55.2 & 62.8  \\ 
        es & 74.9 & 75.4 & 64.6 & 70.7 & 64.4 & 66.1 & 61.1 & 44.9 & 39.7 & 64.6 & 61.0 & 63.8 & 55.0 & 50.2 & 54.3 & 60.7  \\ 
        de & 72.9 & 68.9 & 70.0 & 68.0 & 63.9 & 66.3 & 59.4 & 45.1 & 40.3 & 63.5 & 60.5 & 62.7 & 56.8 & 54.2 & 55.1 & 60.5  \\ 
        fr & 75.9 & 72.2 & 66.0 & 74.3 & 64.2 & 65.3 & 60.6 & 45.5 & 39.7 & 65.2 & 62.0 & 64.5 & 55.4 & 52.0 & 54.7 & 61.2  \\ 
        bg & 69.8 & 66.0 & 61.6 & 64.2 & 69.6 & 66.5 & 59.3 & 45.5 & 38.9 & 60.2 & 60.2 & 60.4 & 54.8 & 50.6 & 53.0 & 58.7  \\ 
        ru & 70.6 & 67.0 & 63.3 & 65.4 & 65.0 & 69.8 & 58.2 & 44.2 & 38.7 & 61.6 & 59.9 & 61.1 & 54.3 & 50.1 & 52.9 & 58.8  \\ 
        el & 64.9 & 63.7 & 59.1 & 62.2 & 60.3 & 60.5 & 67.9 & 45.6 & 40.2 & 59.4 & 59.0 & 56.3 & 53.0 & 49.8 & 52.1 & 56.9  \\ 
        th & 55.0 & 52.9 & 49.8 & 52.0 & 51.5 & 51.7 & 50.8 & 53.5 & 37.9 & 53.2 & 52.5 & 51.1 & 48.4 & 47.1 & 46.0 & 50.2  \\ 
        sw & 54.2 & 52.6 & 48.9 & 50.5 & 49.9 & 49.5 & 49.2 & 42.6 & 50.3 & 49.9 & 50.8 & 49.7 & 47.5 & 47.1 & 47.2 & 49.3  \\ 
        vi & 69.8 & 63.3 & 59.2 & 63.1 & 58.2 & 61.1 & 57.0 & 46.3 & 38.5 & 71.0 & 57.3 & 64.8 & 52.9 & 49.1 & 49.2 & 57.4  \\ 
        ar & 65.7 & 62.6 & 57.6 & 61.4 & 59.2 & 59.9 & 57.1 & 45.5 & 39.2 & 59.3 & 66.3 & 58.0 & 54.1 & 51.5 & 51.9 & 56.6  \\ 
        zh & 69.8 & 62.5 & 58.4 & 61.7 & 57.7 & 59.5 & 53.1 & 43.9 & 38.3 & 63.0 & 57.0 & 70.3 & 50.8 & 48.0 & 49.4 & 56.2  \\ 
        hi & 62.2 & 57.8 & 56.4 & 57.4 & 56.5 & 57.5 & 54.9 & 45.4 & 38.7 & 56.5 & 56.3 & 56.4 & 61.9 & 55.6 & 52.0 & 55.0  \\ 
        ur & 61.0 & 54.6 & 54.3 & 56.6 & 52.9 & 55.1 & 51.2 & 43.6 & 38.9 & 54.0 & 55.3 & 54.3 & 57.2 & 58.9 & 50.1 & 53.2  \\ 
        tr & 62.2 & 57.8 & 56.0 & 56.8 & 56.3 & 55.5 & 54.2 & 43.7 & 39.9 & 55.4 & 55.4 & 55.7 & 54.0 & 51.4 & 62.2 & 54.4  \\ \midrule
        avg. & 67.5 & 63.4 & 59.5 & 62.4 & 59.7 & 60.8 & 57.0 & 45.5 & 40.0 & 60.3 & 58.4 & 59.8 & 54.2 & 51.3 & 52.3 & 56.8 \\ \bottomrule
    \end{tabular}
    \caption{Results for \MODEL on XNLI.}
    \label{tab:gen_salt}
\end{table*}

\begin{table*}[h]
\centering
\small
\setlength{\tabcolsep}{3.5pt}
\begin{tabular}{c|ccccccccccccccc|c}
    \toprule
    & en & es & de & fr & bg & ru & el & th & sw & vi & ar & zh & hi & ur & tr & avg. \\
    \midrule
en & 82.6 & 71.2 & 65.9 & 70.3 & 62.0 & 65.7 & 57.0 & 44.1 & 40.9 & 64.1 & 58.9 & 65.7 & 52.8 & 49.2 & 51.2 & 60.1 \\
es & 74.9 & 75.0 & 65.4 & 71.2 & 63.0 & 65.6 & 59.5 & 44.5 & 40.8 & 62.9 & 60.1 & 62.6 & 52.5 & 48.7 & 51.7 & 59.9 \\
de & 72.6 & 68.0 & 70.5 & 67.4 & 61.7 & 64.9 & 58.0 & 44.4 & 41.4 & 61.0 & 58.8 & 61.4 & 53.6 & 50.4 & 52.2 & 59.1 \\
fr & 74.7 & 71.6 & 65.2 & 74.1 & 62.1 & 64.8 & 58.4 & 44.4 & 40.8 & 62.7 & 59.5 & 62.4 & 52.7 & 48.9 & 51.7 & 59.6 \\
bg & 68.5 & 66.0 & 62.9 & 65.1 & 68.7 & 66.8 & 59.4 & 45.1 & 41.1 & 59.7 & 59.4 & 59.7 & 53.5 & 49.6 & 51.8 & 58.5 \\
ru & 69.9 & 67.1 & 63.5 & 65.9 & 65.0 & 69.5 & 58.2 & 44.7 & 40.9 & 60.8 & 59.2 & 60.6 & 53.1 & 49.5 & 51.5 & 58.6 \\
el & 63.9 & 63.3 & 59.3 & 62.0 & 59.8 & 61.0 & 67.2 & 44.7 & 41.3 & 57.3 & 57.7 & 55.7 & 51.4 & 48.2 & 50.7 & 56.2 \\
th & 56.4 & 54.1 & 51.7 & 53.3 & 51.9 & 52.9 & 51.0 & 50.5 & 40.1 & 52.5 & 51.8 & 51.3 & 48.2 & 46.3 & 46.8 & 50.6 \\
sw & 54.1 & 52.3 & 49.6 & 50.9 & 49.1 & 49.7 & 48.6 & 41.8 & 48.4 & 48.7 & 49.8 & 48.1 & 45.4 & 44.5 & 47.1 & 48.6 \\
vi & 69.9 & 65.0 & 60.5 & 64.1 & 58.6 & 61.8 & 56.0 & 45.1 & 40.4 & 70.5 & 57.4 & 63.4 & 51.5 & 48.0 & 49.1 & 57.4 \\
ar & 64.9 & 62.8 & 58.7 & 61.7 & 58.7 & 60.7 & 56.3 & 44.7 & 41.1 & 58.1 & 65.4 & 57.1 & 52.2 & 49.6 & 50.3 & 56.1 \\
zh & 71.1 & 64.8 & 60.8 & 64.1 & 59.0 & 61.6 & 53.9 & 43.5 & 40.8 & 63.0 & 56.8 & 69.7 & 50.9 & 47.8 & 49.7 & 57.2 \\
hi & 62.2 & 58.9 & 56.7 & 57.9 & 56.5 & 58.3 & 54.3 & 44.5 & 40.8 & 55.3 & 55.8 & 55.3 & 59.8 & 54.2 & 50.4 & 54.7 \\
ur & 61.2 & 56.7 & 56.1 & 57.3 & 54.4 & 57.0 & 53.2 & 43.7 & 40.8 & 54.1 & 56.3 & 54.6 & 56.9 & 57.9 & 49.9 & 54.0 \\
tr & 62.4 & 59.2 & 57.0 & 58.6 & 56.7 & 57.9 & 54.2 & 43.7 & 40.9 & 54.8 & 55.1 & 54.9 & 52.2 & 48.8 & 59.7 & 54.4 \\
    \midrule
avg. & 67.3 & 63.7 & 60.3 & 62.9 & 59.1 & 61.2 & 56.4 & 44.6 & 41.4 & 59.0 & 57.5 & 58.8 & 52.4 & 49.4 & 50.9 & 56.3 \\
    \bottomrule
\end{tabular}
\caption{Results for RS-RP on XNLI.}
\label{tab:gen_rsrp}
\end{table*}

\clearpage
\begin{table*}[h]
\centering
\small
\setlength{\tabcolsep}{3.5pt}
\begin{tabular}{c|ccccccccccccccc|c}
    \toprule
    & en & es & de & fr & bg & ru & el & th & sw & vi & ar & zh & hi & ur & tr & avg. \\
    \midrule
en & 82.3 & 70.3 & 65.8 & 69.7 & 60.5 & 63.1 & 55.3 & 44.6 & 41.1 & 63.9 & 57.7 & 64.6 & 52.0 & 49.5 & 52.3 & 59.5 \\
es & 73.5 & 74.3 & 62.9 & 69.0 & 60.5 & 63.7 & 57.3 & 44.6 & 40.6 & 61.4 & 57.9 & 60.8 & 50.4 & 47.1 & 51.6 & 58.4 \\
de & 71.8 & 65.5 & 70.8 & 65.6 & 59.5 & 63.3 & 55.8 & 44.3 & 41.0 & 60.2 & 56.5 & 60.1 & 52.5 & 49.4 & 52.0 & 57.9 \\
fr & 73.6 & 69.0 & 64.0 & 73.8 & 59.5 & 63.1 & 55.7 & 44.1 & 40.5 & 62.2 & 57.3 & 61.6 & 51.1 & 48.5 & 51.8 & 58.4 \\
bg & 67.8 & 63.7 & 60.8 & 62.5 & 68.2 & 64.2 & 56.0 & 44.2 & 39.9 & 57.4 & 56.3 & 57.8 & 51.2 & 47.2 & 50.3 & 56.5 \\
ru & 69.1 & 65.2 & 62.6 & 64.4 & 62.7 & 68.7 & 55.0 & 44.2 & 39.9 & 59.0 & 56.7 & 58.6 & 50.6 & 46.8 & 50.0 & 56.9 \\
el & 62.7 & 61.4 & 58.0 & 60.2 & 57.1 & 57.7 & 66.4 & 44.4 & 40.5 & 56.4 & 55.6 & 54.0 & 49.6 & 46.8 & 50.7 & 54.8 \\
th & 54.8 & 52.0 & 49.9 & 51.3 & 49.1 & 50.4 & 49.0 & 53.0 & 39.4 & 51.1 & 49.9 & 49.3 & 45.9 & 44.8 & 45.4 & 49.0 \\
sw & 54.2 & 51.2 & 48.7 & 50.5 & 47.2 & 47.9 & 47.9 & 41.8 & 50.0 & 48.5 & 49.1 & 48.5 & 45.4 & 44.4 & 45.8 & 48.1 \\
vi & 67.4 & 60.3 & 57.4 & 61.2 & 52.9 & 57.1 & 52.9 & 44.2 & 39.8 & 70.3 & 53.3 & 62.0 & 49.2 & 45.9 & 47.5 & 54.8 \\
ar & 63.9 & 60.4 & 57.0 & 59.5 & 54.5 & 57.1 & 53.3 & 43.9 & 40.4 & 55.4 & 64.8 & 55.2 & 50.3 & 48.4 & 49.9 & 54.3 \\
zh & 67.9 & 59.9 & 57.2 & 59.9 & 53.4 & 56.5 & 50.4 & 42.7 & 39.6 & 60.8 & 53.5 & 69.2 & 48.0 & 45.7 & 48.0 & 54.2 \\
hi & 61.4 & 55.5 & 55.0 & 55.3 & 52.6 & 54.4 & 51.9 & 43.8 & 40.3 & 53.8 & 53.1 & 53.7 & 59.7 & 52.7 & 49.9 & 52.9 \\
ur & 60.1 & 54.0 & 53.9 & 55.1 & 48.8 & 51.5 & 49.6 & 41.9 & 39.7 & 50.0 & 52.1 & 52.3 & 54.4 & 57.7 & 48.2 & 51.3 \\
tr & 61.0 & 55.1 & 53.6 & 55.1 & 52.0 & 52.6 & 50.9 & 42.4 & 40.7 & 52.3 & 52.0 & 53.2 & 49.7 & 47.3 & 60.9 & 51.9 \\
    \midrule
avg. & 66.1 & 61.2 & 58.5 & 60.9 & 55.9 & 58.1 & 53.8 & 44.3 & 40.9 & 57.5 & 55.1 & 57.4 & 50.7 & 48.1 & 50.3 & 54.6 \\
    \bottomrule
\end{tabular}
\caption{Results for mBERT on XNLI.}
\label{tab:gen_mbert}
\end{table*}

\end{document}